\pdfoutput=1

\documentclass[11pt]{article}
\usepackage[dvipsnames]{xcolor}

\usepackage{acl}

\usepackage{amsmath}
\usepackage{amssymb}
\usepackage{makecell}
\usepackage{times}
\usepackage{latexsym}
\usepackage{booktabs}
\usepackage[T1]{fontenc}

\usepackage[utf8]{inputenc}
\usepackage{comment}

\usepackage{microtype}
\usepackage{mathtools}
\usepackage{float}
\usepackage{bm}
\usepackage[font=small]{subfig}
\usepackage{cleveref}

\crefname{section}{\S}{\S\S}
\Crefname{section}{\S}{\S\S}
\crefname{table}{Table}{}
\crefname{figure}{Figure}{}
\crefname{algorithm}{Algorithm}{}
\crefname{equation}{eq.}{}
\crefname{appendix}{App.}{}
\crefname{prop}{Proposition}{}
\crefformat{section}{\S#2#1#3}
\usepackage{enumitem}

\usepackage{todonotes}

\renewcommand\cite{\citep}	
\newcommand{\vs}{\mathbf{s}}

\newcommand{\vz}{\mathbf{z}}
\newcommand{\vp}{\mathbf{p}}

\newcommand{\vh}{\mathbf{h}}

\newcommand{\MLP}{\mathrm{MLP}}

\newcommand{\calT}{\mathcal{T}}

\newcommand{\calD}{\mathcal{D}}

\newcommand{\saveForCR}[1]{}

\newcommand{\calM}{\mathcal{M}}

\newcommand{\calC}{\mathcal{C}}

\newcommand{\calW}{\mathcal{W}}
\newcommand{\calS}{\mathcal{S}}
\newcommand{\calL}{\mathcal{L}}
\newcommand{\real}{\mathbb{R}}

\definecolor{purple}{HTML}{CC00CC}
\definecolor{red}{HTML}{FF3333}
\definecolor{blue}{HTML}{0070E0}

\newcommand{\smdcolour}[1]{\textcolor{purple}{#1}}
\newcommand{\wsdcolour}[1]{\textcolor{red}{#1}}
\newcommand{\mpdcolour}[1]{\textcolor{blue}{#1}}
\newcommand{\black}[1]{\textcolor{black}{#1}}

\newcommand{\vtheta}{{\boldsymbol \theta}}
\newcommand{\vphi}{{\boldsymbol \phi}}
\newcommand{\vpsi}{{\boldsymbol \psi}}

\newcommand{\pwn}{p_\vtheta}
\newcommand{\pwsd}{p_\vphi}
\newcommand{\pjoint}{p_{\langle \vtheta, \vphi \rangle}}
\everypar{\looseness=-1}
\newcommand{\jointloss}{\calL}
\newcommand{\metloss}{\calL_\text{SMD}}
\newcommand{\wsdloss}{\calL_\text{WSD}}

\newcommand{\defn}[1]{\textsl{{#1}}}

\newcommand{\word}[1]{\textit{#1}}

\newcommand{\sense}[2]{\word{#1}$_#2$}

\newcommand{\tabspace}{\addlinespace[0.25em]}
\newcommand{\halftabspace}{\addlinespace[0.1em]}

\DeclareMathOperator*{\mean}{mean}

\usepackage{multirow}

\usepackage{pifont}
\newcommand{\cmark}{{\ding{51}}}%
\newcommand{\xmark}{{\ding{55}}}%

\title{Metaphorical Polysemy Detection:\\Conventional Metaphor meets Word Sense Disambiguation}

\author{Rowan Hall Maudslay$^{1,2}$ \hspace{1em} Simone Teufel$^{1}$\\
${}^1$Dept.\ of Computer Science \& Technology \hspace{.5em} ${}^2$Magdalene College \\
University of Cambridge \\
  {\tt \{rh635,sht25\}@cam.ac.uk}}

\begin{document}
\maketitle
\begin{abstract}
Linguists distinguish between novel and conventional metaphor, a distinction which the metaphor detection task in NLP does not take into account. Instead, metaphoricity is formulated as a property of a token in a sentence, regardless of metaphor type. In this paper, we investigate the limitations of treating conventional metaphors in this way, and advocate for an alternative which we name \defn{metaphorical polysemy detection} (MPD). In MPD, only conventional metaphoricity is treated, and it is formulated as a property of word senses in a lexicon. We develop the first MPD model, which learns to identify conventional metaphors in the English WordNet. To train it, we present a novel training procedure that combines metaphor detection with \defn{word sense disambiguation} (WSD). For evaluation, we manually annotate metaphor in two subsets of WordNet. Our model significantly outperforms a strong baseline based on a state-of-the-art metaphor detection model, attaining an ROC-AUC score of $.78$ (compared to $.65$) on one of the sets. Additionally, when paired with a WSD model, our approach outperforms a state-of-the-art metaphor detection model at identifying conventional metaphors in text ($.659$ F1 compared to $.626$).
\end{abstract}

\section{Introduction}\label{sec:intro}

Linguists differentiate between two types of metaphor: novel and conventional. 
While novel metaphors are creative expressions made in a particular situation by one particular individual, conventional metaphors are those which have been widely adopted by a language community. 
Consider:
\begin{enumerate}[label={(\arabic*)}]
\itemsep0em 
\item The \word{attack} began at dawn. \label{ex:literal}
\item The government has come under \word{attack}. \label{ex:conventional}
\item The government \word{torpedoed} the housing bill. \label{ex:novel}
\end{enumerate}
Example~\ref{ex:literal} is a literal usage of \word{attack} which refers to a military offensive. 
Example~\ref{ex:conventional} is a conventional metaphor referring to intense verbal criticism. 
Example~\ref{ex:novel} is a novel metaphor, which reuses the well-established imagery from \ref{ex:conventional}, but extends it with the word \word{torpedoed}.

\begin{figure}
    \centering
    ~~\includegraphics[width=0.9\columnwidth]{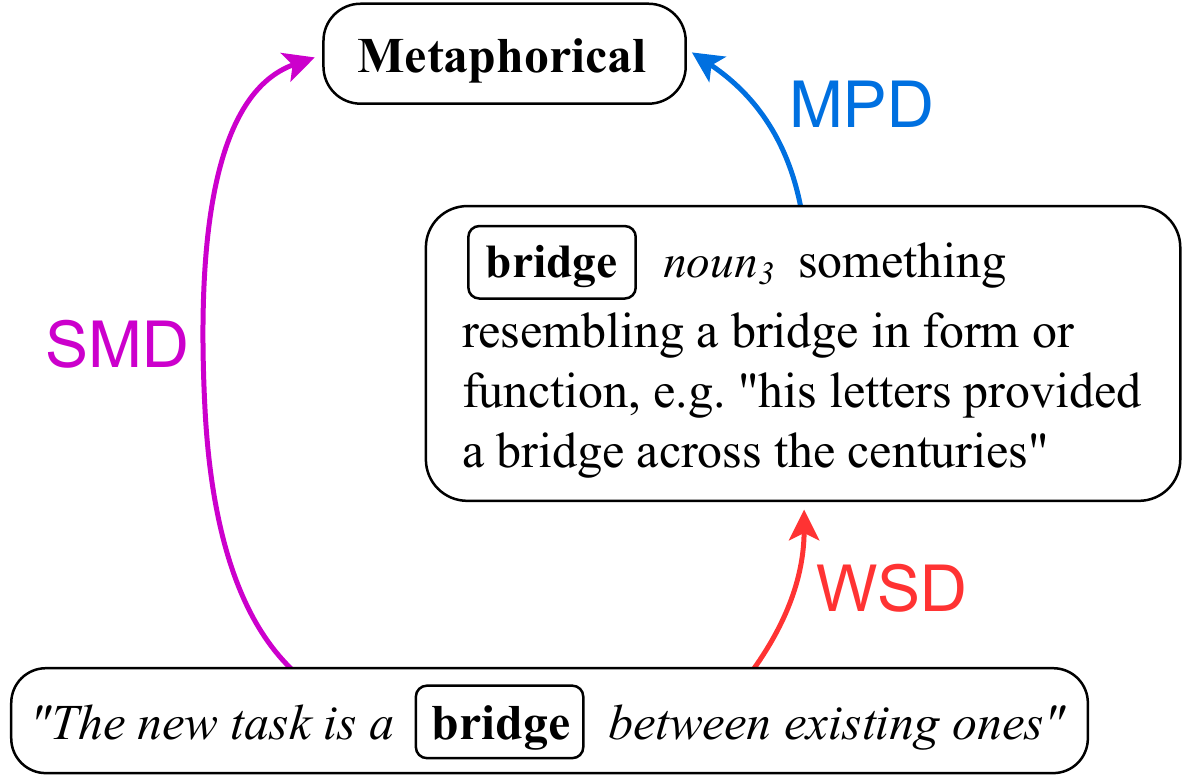}
    \caption{\mpdcolour{Metaphorical polysemy detection} is the missing link between \smdcolour{standard metaphor detection} and \wsdcolour{word sense disambiguation}}
    \label{fig:overview}
\end{figure}

Detecting novel metaphors and conventional metaphors are fundamentally different tasks. 
Novel metaphors are creative usages of words, which violate statistical patterns of language \citep[e.g.][]{wilks1975preferential, wilks1978preferences}, and which extend the meaning of a word into unexpected areas of semantic space. 
Conventional metaphors, on the other hand, can be considered lexicalised: examples~\ref{ex:literal} and~\ref{ex:conventional} come from the gloss of two senses of \word{attack} in WordNet \citep{miller1995wordnet}, where the novel sense of \word{torpedoed} in example~\ref{ex:novel} is not yet captured. 
Despite these inherent differences, in NLP the standard metaphor detection task, abbreviated as SMD here, makes no distinction between novel and conventional metaphor.

\begin{table*}[ht!]
    \centering
    \begin{tabular}{l l c} \toprule
        \textbf{Sense} & \textbf{Definition} & \textbf{Metaphor?} \\ \midrule
        \makecell{\sense{adopt}{1}} & \makecell[l]{take into one's family (e.g.\ \textit{They adopted two children} \\ \textit{from Nicaragua})} & \makecell{\xmark} \\ \tabspace
         \makecell{\sense{adopt}{2}} & \makecell[l]{choose and follow; as of theories, ideas, policies, strategies \\ or plans} & \makecell{\cmark} \\ \tabspace
         \makecell{\sense{adopt}{3}} & \makecell[l]{take on a certain form, attribute, or aspect (e.g.\ \textit{he adopted} \\ \textit{an air of superiority})} & \makecell{\cmark} \\ \tabspace
         \makecell{\sense{adopt}{4}} & \makecell[l]{take up the cause, ideology, practice, method, of someone and \\ use it as one's own (e.g.\ \textit{They adopted the Jewish faith})} & \makecell{\cmark} \\

       \bottomrule
    \end{tabular}
    \caption{Verbal senses of \word{adopted} in WordNet}
    \label{tab:wordnet_met}
\end{table*}

Although recent works have produced resources which distinguish between novel and conventional metaphors \citep{parde-nielsen-2018-corpus, do-dinh-etal-2018-weeding}, both metaphor types are annotated as properties of tokens, 
suitable for SMD.
In this work, we demonstrate the shortcomings of this formulation, and argue that conventional metaphoricity is best treated not as a property of word occurrences in a sentence, but of word senses in a lexicon. 
With this in mind, we investigate the problem of assigning metaphoricity ratings to word senses, a problem we name metaphorical polysemy detection (MPD). 
We build the first model of MPD, which identifies metaphorical senses in WordNet \citep{miller1995wordnet}.

No training data is available for this task. We design a novel training regime which utilises existing resources, which works by decomposing SMD into two steps: word sense disambiguation (WSD) and MPD (see \cref{fig:overview}). 
More specifically, we pair our MPD model with a state-of-the-art WSD model, and train them in conjunction on SMD data, treating word sense as a latent variable.

To investigate the performance of our model, we establish an evaluation framework for MPD. To collect test data, we perform an annotation study, and label two subsets of WordNet for metaphoricity (${\kappa=0.78}$). Metaphor detection is typically evaluated using F1-score, which measures how well a model can judge metaphoricity in absolute terms. For MPD, we additionally introduce a new quantity for evaluation, which we call \defn{relative metaphoricity}. It measures whether a model is able to correctly identify whether one sense is more metaphorical than another, even if it is unable to correctly determine where they sit around a threshold. It is calculated using ROC-AUC.

On one of our test sets, consisting of words from a large resource of conventional metaphors \citep[the Master Metaphor List;][]{lakoff1994master}, our MPD model attains $.78$ ROC-AUC and $.60$ F1, significantly higher than a strong baseline which uses a state-of-the-art SMD model \citep[MelBERT;][]{choi-etal-2021-melbert}, which scores $.65$ and $.54$ respectively. 
Additionally, for SMD on conventional metaphors, when our model is paired with a WSD model it attains $.659$ F1, significantly better than the state-of-the-art SMD model ($.626$).

\section{Metaphor in the Lexicon}

A fundamental feature of language is \defn{polysemy}, the phenomenon of a wordform exhibiting multiple meanings which are systematically related. 
When meanings are related by metaphorical similarity, the phenomenon is called \defn{metaphorical polysemy}. 
An example is the word \word{attack}, which describes either a physical confrontation, or---metaphorically---heated criticism (e.g.\ ``she \word{attacked} my arguments''). 
Instances of metaphorical polysemy are called \defn{conventional metaphors}.

In this section, we describe how conventional metaphors materialise in lexica (\cref{sec:metwordsense}). With this in mind, we illustrate the shortcomings of the existing metaphor detection task (\cref{sec:md_problems}), and propose an alternative formulation which is more suitable for conventional metaphors (\cref{sec:mpd_solutions}).

\subsection{Metaphoricity of Word Senses} \label{sec:metwordsense}

In a lexicon, metaphoricity can be formulated as a property of word senses.
Consider the example definitions of the word \word{adopted} in WordNet \citep{miller1995wordnet}, which are shown in \cref{tab:wordnet_met} (some senses are excluded for brevity). 
Sense \sense{adopt}{1} is literal, but the other senses are conventional metaphors.

The conventionality of a metaphor lends itself to treatment as a continuous property. For instance, a metaphor used by an entire language community would be highly conventionalised, while a metaphor used by speakers in a particular geographic region would be moderately conventionalised. 
\defn{Novel metaphors} sit at the extreme other end of the spectrum: they are creative expressions with ad-hoc meanings.\footnote{Novel metaphors can be further subdivided into metaphors which are novel lexicalisations of conventional imagery, as with example~\ref{ex:novel} above, and metaphors which create new imagery.} By definition, lexica designed to encode widespread language use will only encode the most conventional of metaphors as word senses.

In order to formulate metaphoricity as a property of word senses, a lexicon needs to encode the literal and metaphorical meanings of a wordform as separate senses.
The criteria that need to be met for two meanings to be considered distinct is highly contentious, with some lexicographers arguing that senses should be represented as spectra rather than discrete units \citep[see][Ch.\ 3]{cruse1986lexical}. 
In practice, however, lexica describe senses as discrete objects. Moreover, in most lexica the granularity of these senses is high enough to separate metaphorical senses from literal ones, which thus matches our criteria. That being said, no existing lexicon systematically labels senses for metaphoricity (the decision of whether to label metaphorical senses as such is left to the lexicographer's discretion).

\begin{table*}[t!]
    \centering
    \begin{tabular}{lc} \toprule
         \textbf{Sentence Excerpt} & \textbf{Prediction} \\ \midrule
         \textit{...the South Carolina nullifiers \textbf{adopted} the principle of state interposition...} & \textcolor{black!30!green}{Metaphorical} \\ \halftabspace
         \textit{...the methods he \textbf{adopted} to accomplish...} & \textcolor{black!30!green}{Metaphorical} \\ \halftabspace
         \textit{...the denomination's 16 basic beliefs \textbf{adopted} in 1966...} & \textcolor{black!30!green}{Metaphorical} \\ \halftabspace
         \textit{...the Government's new feed grain program was \textbf{adopted}; the program...} & \textcolor{red}{Literal} \\ \halftabspace
         \textit{...the Albany Plan of Union, which, had it been \textbf{adopted}, might...} & \textcolor{red}{Literal} \\ \halftabspace
         \textit{...the use of target-hunting noses on the projectiles has been \textbf{adopted}, and...} & \textcolor{red}{Literal} \\
         \bottomrule
    \end{tabular}
    \caption{MelBERT makes inconsistent predictions for occurrences of \word{adopted} with sense \sense{adopt}{2} in SemCor}
    \label{tab:melbert_preds}
\end{table*}

If word senses in a lexicon like WordNet were labeled for metaphoricity, this would be useful in a number of areas of study. 
When a language community need to reference a newly-arisen meaning, instead of creating a new wordform, often an existing word is extended metaphorically (e.g.\ a computer \word{mouse}). This process of metaphorical extension is a cornerstone of lexical-semantic language change \citep{koch2016meaning}, and knowing which word senses were metaphorical would open up new possibilities to study this phenomenon.
The metaphors which do enter the lexicon participate in sophisticated patterns, preserving the structure of their literal domains: just as one can \word{attack} an opponent in a debate, so too can claims be \word{defended}, different rhetorical \word{strategies} adopted, and so on. 
These patterns are known as \defn{conceptual metaphors} \citep{lakoff1980metaphors}, and knowing which word senses in a lexicon were metaphorical would create new opportunities to analyse their substructures \citep[this has been previously recognised by][]{lonneker2003way}. 

\subsection{Issues with Standard Metaphor Detection} \label{sec:md_problems}

In the \defn{standard metaphor detection} (SMD) task, metaphoricity is formulated as a property of a token in a sentence. 
The dataset which is almost universally used for this is the VUAMC \citep{steen2010method}.
It contains tokenised English sentences, in which words are annotated with binary metaphoricity labels.
It was the subject of two recent metaphor detection shared tasks \citep{leong-etal-2018-report, leong-etal-2020-report}. 

The VUAMC was annotated following an adapted version of the Metaphor Annotation Procedure \citep[MIP;][]{semino2007mip}. 
To annotate each token, MIP involves a three-stage process:
\begin{enumerate}
    \item Establish the contextual meaning of the word, based on the other words in the sentence. \label{mip1}
    \item Determine whether the word has a more basic meaning which occurs in different contexts (where ``basic'' meanings are defined as those more concrete, related to bodily action, less vague, and/or historically older). \label{mip2}
    \item If the word does have a more basic meaning in different contexts, decide whether the meaning in this context can be understood in comparison with the more basic meaning. \label{mip3}
\end{enumerate}
If the wordform does have a more basic meaning which the contextual meaning relates to, the token is labeled as a metaphor. 

To perform well in SMD, a model must emulate this procedure.
There are two drawbacks to this task formulation:

\paragraph{Learning WSD Implicitly} For conventional metaphors, stage~\ref{mip1} of MIP is very similar to \defn{word sense disambiguation} (WSD). In WSD, the goal is to identify the sense evoked by a wordform in a sentence. This is a hard task in its own right, and, in order to perform well, models of SMD are expected to learn to do this implicitly. 

We can check whether an SMD model has learnt to do this by analysing its predictions for wordforms which appear in multiple distinct contexts which all evoke the same sense. Tokens which evoke the same sense should always receive the same metaphoricity prediction (either they should all be metaphorical, or all literal).
This is because, for conventional metaphors, metaphoricity is a property of a word sense (\cref{sec:metwordsense}). 
After the sense of a token is established, stages~\ref{mip2} and~\ref{mip3} of MIP should always yield the same prediction; if a model makes inconsistent predictions for different invocations of the same sense, that suggests that it is incorrectly establishing the contextual meaning of the token.

To investigate whether a state-of-the-art model is able to adequately perform stage~\ref{mip1} of MIP, we perform an error analysis of MelBERT \citep{choi-etal-2021-melbert}.
More specifically, we compute MelBERT's metaphoricity predictions for every token in an English sense-tagged corpus \citep[SemCor;][]{miller1994semcor}, then extract all the predictions of the same word sense, which should always elicit a consistent prediction. \cref{tab:melbert_preds} shows MelBERT's predictions for different instances of the word \word{adopted} which all evoke the same metaphorical sense, \sense{adopt}{2}.\footnote{Data here comes from our re-implementation of MelBERT trained on the `all' subset, described in \cref{sec:setup}.} MelBERT misclassifies the bottom three examples as literal. 
In general, for word senses which occur multiple times in the data, 
MelBERT gave contradictory classifications $31.8\%$ of the time. 
It becomes more likely that a system will make an inconsistent classification when there are more occurrences of a word sense in the data: for word senses which occurred $15$ or more times, the misclassification rate was $49.3\%$.

This error analysis suggests that MelBERT is not performing stage~\ref{mip1} of MIP adequately. 
The errors in \cref{tab:melbert_preds} result not from a failure to reason about metaphoricity, but from MelBERT incorrectly establishing the contextual meaning of \word{adopted} in the first place. We argue that this is beyond the remit of metaphor detection, since it is a hard challenge in its own right, covered by other NLP tasks.

\paragraph{Word's Senses Undefined} 
After stage~\ref{mip1} of MIP has been performed, stages~\ref{mip2} and~\ref{mip3} require information about how the contextual meaning of a wordform relates to the other senses it may take. 
More specifically, models need to determine the most basic sense of the wordform, and relate that sense to the contextual meaning.
However, the VUAMC does not provide information about other meanings a wordform could assume.
Because of this,  model architects typically assume that static word embeddings will represent the basic sense of a wordform \citep[e.g.][]{choi-etal-2021-melbert, mao-etal-2018-word, zayed-etal-2018-phrase, wu-etal-2018-neural}. 
We argue that this assumption is problematic, because static word embeddings do not distinguish between word senses, and are sensitive to occurrence frequency. 
If one sense is much more frequent than others, static word embeddings will primarily encode that sense---but the basic sense of a wordform is often not the most frequent.

Fortunately, we know that the most basic sense of a wordform will be one of its senses in a lexicon.
If we model metaphoricity as a property of word senses, we create scope for researchers to explicitly investigate the relations between different senses, which we argue is the core challenge posed by metaphor.

\subsection{A New Task Formulation for Conventional Metaphor}
\label{sec:mpd_solutions}

Let $w\in\calW$ be all the wordforms in a language. To represent metaphoricity, let $\calM$ be the set $\{0, 1\}$ with members~$m$, such that $1$ represents metaphorical and $0$ represents literal. 

A token ${t\in\calT}$ is an occurrence of a wordform in a sentence.
Using $\circ$ to denote concatenation, we define a token as a wordform~$w$ surrounded by a prefix and suffix string ($\vp$ and 
$\vs$)
 in a sentence:
\begin{align} \label{eq:def_context}
    \calT = \{\langle \vp, w, \vs \rangle \mid \vp \circ w \circ \vs \in \calW^* \}
\end{align}
In SMD, the goal is to determine whether a token is metaphorical, i.e.\ to construct a model of ${p(m \mid t)}$.

A word sense is a particular meaning of a word, which are typically represented in lexica as definitions. Let~$\calD$ be the complete set of definitions in a lexicon (in WordNet these correspond to synsets), with elements~$d$. 
A lexicon maps a wordform~$w$ to a subset of the definitions ${\calD_w\subseteq\calD}$.\footnote{Formally, if $\mathbb{P}(x)$ denotes the powerset of $x$, then a lexicon is a map~${\calW\rightarrow\mathbb{P}(\calD)}$.}
Some lexica (WordNet included) associate multiple synonymous wordforms with the same definition; because of this, we define a word sense~$s\in\calS$ as a tuple consisting of a wordform and a definition:
\begin{align} \label{eq:def_senses}
    \calS = \{\langle w, d \rangle \mid w \in \calW, d \in \calD_w \}
\end{align}

Because conventional metaphors are lexicalised, we propose that the task of detecting conventional metaphors should be formulated as modelling the metaphoricity of word senses, i.e.\ constructing models of ${p(m\mid s)}$. We call this \defn{metaphorical polysemy detection} (MPD). 
This formulation of conventional metaphoricity alleviates the problems with SMD outlined in \cref{sec:md_problems}.

\section{Learning Conventional Metaphor} \label{sec:model}

In this section, we describe the first MPD model (\cref{sec:wn_met_pred}). To train it, it is embedded into an SMD pipeline (\cref{sec:model_overview}).

\subsection{An Architecture for MPD}\label{sec:wn_met_pred}

To model MPD (see \cref{sec:mpd_solutions}), we implement ${p(m\mid s)}$ as a multi-layer perceptron (MLP). In \cref{eq:def_senses}, we defined a word sense~$s$ as a tuple consisting of a wordform~$w$ and a definition~$d$. Our MPD model's MLP is input with a pair of embeddings, for $w$ and $d$. Suppose we have a $k$-dimensional embeddings space for words~${w\in\calW}$, and a $k$-dimensional embedding space for definitions~${d\in\calD}$. 
Let us define two functions to map inputs into these spaces, ${\texttt{TypeEmb} : \calW \rightarrow {\real} ^{k}}$ and 
${\texttt{SynsetEmb} : \calD \rightarrow {\real} ^{k}}$.\footnote{For  $\texttt{SynsetEmb}$ we use ARES embeddings \citep{scarlini-etal-2020-contexts}, and for $\texttt{TypeEmb}$ we use BERT embeddings of wordforms \citep[following][]{choi-etal-2021-melbert}; see \cref{app:training}.} We concatenate these embeddings, then pass them through an MLP and a sigmoid to get a probability distribution. Letting 
$\texttt{MLP}_\vpsi^{\langle u,v \rangle} : \real^u \rightarrow \real^v$ be an MLP parameterised by $\vpsi$ with an input size $u$ and output size $v$, we have
\begin{align}
& \vh = \texttt{TypeEmb}(w)\circ \texttt{SynsetEmb}(d) \nonumber \\
& \pwn (m\mid s) = \sigma \big(\texttt{MLP}^{\langle 2k, 1 \rangle}_\vtheta \big( \vh \big)\big)
\end{align}

\subsection{Teasing MPD out of SMD using WSD}\label{sec:model_overview}

Training this model would be trivial if we had $\langle m,s \rangle$ tuples, but no training data of this form currently exists. 
In WSD, the task is to produce models of ${p(s\mid t)}$ for a corpus ${\langle s, t \rangle~\in~\calC_\text{WSD}}$. In SMD, on the other hand, the task is to produce models of  ${p(m\mid t)}$, using a corpus ${\langle m,t \rangle~\in~\calC_\text{SMD}}$. No corpus contains both~$m$ and $s$ annotations in parallel.
In this section, we describe a training procedure that works by pairing an MPD model with a WSD model, and training them in conjunction on SMD, thus making it possible to use existing resources of $\calC_\text{WSD}$ and $\calC_\text{SMD}$ to train MPD. 
We do not believe that this procedure is an optimal way to learn MPD, as errors in the WSD component will degrade MPD performance. 
However, given that these are the only resources currently available, 
we design a method to use them for MPD, and in doing so empirically test whether they can result in a sufficiently good MPD model.

By marginalising out word sense, we can decompose SMD into two parts, introducing sense~$s$ as a latent variable.
The first part disambiguates senses (WSD), and the second predicts metaphoricity based on the sense (MPD):
\begin{align} \label{eq:joint}
    \smdcolour{\underbrace{\black{\vphantom{\sum}\pjoint (m\mid t)}}_{\substack{\text{SMD}}}} = 
     \sum_{s\in \calS} \mpdcolour{\underbrace{\black{\vphantom{\sum}\pwn (m\mid s)}}_{\text{MPD}}} ~\wsdcolour{\underbrace{\black{ \vphantom{\sum}\pwsd(s\mid t)}}_{\text{WSD}}} 
\end{align}
where $\vtheta$ and $\vphi$ are disjoint sets of parameters (colours correspond with \cref{fig:overview}). 
Introducing WSD into SMD like this makes explicit what SMD models typically attempt to learn implicitly (see \cref{sec:md_problems}).
The WSD component outputs a probability distribution over the senses~${s\in\calS}$, and the MPD component outputs the probability that each $s$ is metaphorical.\footnote{For a particular token ${t=\langle \vp, w, \vs \rangle}$ and sense ${s=\langle w', d \rangle}$, the probability ${p(s\mid t)}$ will only be non-zero if ${w=w'}$; in practice, this can be used to reduce the computation.} 
This formulation assumes that metaphoricity is conditionally independent of token given sense; we assume that the context of a word's usage tells us its sense, and that that alone is enough to predict metaphoricity.

Taken as a whole, the combined model in \cref{eq:joint} can be trained end-to-end by minimising its cross-entropy on an SMD dataset, 
\begin{align}\label{eq:met_loss}
    \metloss(\vtheta, \vphi) = -\sum_{\mathclap{\substack{
    \langle m, t \rangle  \\ \in 
    \calC_\text{SMD}}}} \frac{\log \pjoint (m\mid t)}{|\calC_\text{SMD}|}
\end{align}
However, this approach to training would leave the model to infer $s$ implicitly. 
Instead, we can complement this objective with another, which trains the WSD model in isolation on another dataset ($\calC_\text{WSD}$), also using cross-entropy,
\begin{align}\label{eq:wsd_loss}
    \wsdloss(&\vphi) = - \sum_{\mathclap{\substack{
    \langle s, t \rangle \\ \in 
    \calC_\text{WSD}}}} \frac{ \log  \pwsd (s\mid t) }{|\calC_\text{WSD}|}
\end{align}
These two objectives can be combined into a multi-task objective, using $\alpha$ as a hyperparameter which regulates the ratio between them, yielding the final global loss function
\begin{alignat}{3}\label{eq:joint_loss}
    \jointloss(\vtheta, \vphi) =& &&~~\alpha & & \cdot \metloss(\vtheta, \vphi)~+ \\
    & (1 &&- \alpha) & & \cdot \wsdloss(\vphi) \nonumber
\end{alignat}
Any supervised WSD architecture can be used in conjunction with our MPD model.

\section{Evaluation}\label{sec:evaluation}

In this section, we describe the evaluation data we collected for MPD (\cref{sec:eval_sets}), and introduce relative metaphoricity, an aspect of metaphor which it is only possible to evaluate with MPD (\cref{sec:relative_met}). 
We then describe our experimental setup (\cref{sec:setup}) and results (\cref{sec:results}).

\subsection{Evaluation Data for MPD} \label{sec:eval_sets} \label{sec:wn_pilot}

Our approach to training (\cref{sec:model_overview}) alleviates the need for labeled MPD training data. 
However, we still need evaluation data.
For this we collect annotation. 
In this section, we describe the main evaluation set we collected. This evaluation set consists of data from the Master Metaphor List, which is the largest list of conventional metaphors \citep[collated by][]{lakoff1994master}.

\paragraph{Data for Annotation} 
We sample $250$ commonly cited metaphor examples from the MML, filtering out wordforms which have more than $10$ or fewer than $3$ senses. We annotate the remaining $98$ wordforms for metaphoricity (a total of $554$ word senses). 

\paragraph{Annotation Procedure} 
We adapt MIP stages~\ref{mip2} and~\ref{mip3} for the annotation task (our guidelines are presented in \cref{sec:guidelines}). 
For MPD, we require a lexicon which consistently separates literal and metaphorical senses (see \cref{sec:metwordsense}). We use the annotation phase to also check whether WordNet complies with this. 
Along with options for ``metaphorical'' and ``literal'', annotators also have the option to label senses as ``mixed'', meaning that the sense combined literal and metaphorical definitions.
Our data is annotated by two judges, who are native speakers of English. 

\paragraph{Analysis} 
At least one of the annotators labeled a sense as ``mixed'' $5.4\%$ of the time.
We conclude from this that although WordNet mixes some literal and metaphorical senses, for the most part the granularity of the sense inventory was high enough to separate them. 
Of those which neither thought was mixed, inter-annotator agreement was measured at ${\kappa=0.78}$ ($N{=}539, n{=}2, k{=}2$): where the senses were clearly distinct, annotation was consistent. 
Of the senses that the judges agreed upon, $52\%$ were metaphorical and $48\%$ were literal.

\subsection{Relative Metaphoricity} \label{sec:relative_met}

Metaphor detection performance is typically reported using F1-score, which factors in the precision and recall of a classifier. 
This evaluates a model's ability to judge whether the metaphoricity level of an input passes a threshold to be considered metaphorical, in an absolute sense.

With absolute metaphoricity judgments, there are lots of edge cases, which do not cleanly fit into a metaphorical--literal binary. 
For example, while a physical \word{attack} (e.g.\ an outlaw with a knife) is widely accepted as being literal, and an \word{attack} with words (e.g.\ in the perpetrator's court hearing) is widely accepted as being metaphorical, the case is less clear-cut for a sporting \word{attack} (e.g.\ when they are playing football with other inmates): here we have a physical act which can result in injury, but that takes place within the confines of a game, and certainly does not involve weapons.\footnote{\textit{Ad extremum}, this line of reasoning might lead us to ask whether a sporting \word{attack} in rugby (or even boxing) would be more literal than it would be in non-contact sports.}
It is difficult to decide whether an edge case like this is metaphorical or not, if those are the only options. 
Most people would likely agree, however, that a sporting attack is \emph{more} metaphorical than an attack with a knife, and \emph{less} metaphorical than an attack within a debate. 

We are not the first to note these flaws in absolute metaphoricity judgements; they have previously been used to motivate proposals for the treatment of metaphor on graded or continuous scales 
\citep[e.g.][]{dunn-2014-measuring, mohler-etal-2016-introducing}. Even though MPD is a binary classification task, it additionally gives us the opportunity to evaluate whether a model can judge which of a word's senses are more metaphorical than others, even if it gets the exact threshold wrong. 
We call this a measure of \defn{relative metaphoricity}. 

Beneath the surface, models assign a probability of metaphoricity, rather than an absolute judgement. 
With MPD, we can compute a model's predictions for all the senses of a word, and investigate whether it puts these probabilities in the right order, even if it gets absolute predictions ``wrong'' in an absolute sense.
A metric for this is described in \cref{sec:setup}.

\subsection{Experimental Setup}\label{sec:setup}

In this section, we describe the models and data we experimented with. Additional details can be found in the appendices.

\paragraph{MPD Models} We compare our MPD model (\textbf{Ours}) to several baselines. To establish a lower-bound of performance for MPD, we compute two baselines which make predictions randomly or by choosing the most common class (\textbf{Random} and \textbf{Majority}). In addition to these, we compute a strong baseline: using a state-of-the-art SMD model \citep[MelBERT;][]{choi-etal-2021-melbert}, we compute the metaphoricity probability of all occurrences of $s$ in WSD data, and take the average (\textbf{MelBERT Average}). More formally, let $t \in \calC_\text{WSD}^{s}$ be every token in the WSD data which elicits a specific word sense~$s$.
We compute the average metaphoricity probability of that word sense by taking the mean of MelBERT's metaphoricity prediction for all tokens that evoke that sense: 
\begin{equation}
    p(m\mid s) = \mean_{t \in \calC_\text{WSD}^{s}} 
    \underbrace{p(m \mid t)}_{\substack{\text{MelBERT}}} 
\end{equation}
If $|\calC_\text{WSD}^{s}|=0$ (i.e.\ there are no occurrences of  $s$ in the WSD data) then we default to the random baseline (and take the probability as random).

\paragraph{WSD Models} We experiment with two WSD models: a baseline (\textbf{BERT WSD Baseline}) and a state-of-the-art approach \citep[\textbf{EWISER};][]{bevilacqua-navigli-2020-breaking}. Additional description is left to \cref{sec:wsd_models}.

\paragraph{SMD Models}
We train the combined model (\textbf{WSD+MPD}) from \cref{eq:joint}, using our MPD model and both WSD models (we select the highest performing variant). We compare the overall SMD performance of this combined model to two other approaches, a baseline (\textbf{BERT SMD Baseline}) and a high-performing model (\textbf{MelBERT}). Additional description is left to \cref{app:smd_models}.

\paragraph{MPD Data}
To evaluate MPD performance, we use the evaluation set described in \cref{sec:eval_sets} (\textbf{MML}).
We also collected a second set using the same procedure, designed to evaluate model generalisation ability to words which are out-of-vocabulary of the training data (\textbf{OOV}). We randomly sampled $100$ wordform from the WordNet vocabulary which are not included in the vocabularies of $\calC_\text{SMD}$ or $\calC_\text{WSD}$, and asked one of the annotators from \cref{sec:wn_pilot} to follow the same procedure as before. The final evaluation sets for MML and OOV consist of $535$ and $492$ word senses respectively, after senses which the annotator labeled as ``mixed'' were removed.
As an additional third set, we use the data of \citet{mohammad-etal-2016-metaphor}, who annotated verbs in WordNet for metaphoricity (\textbf{Verbs}). 
Their data consists of $1{,}679$ word sense annotations, covering $440$ verbal wordforms.

\paragraph{WSD Data} For $\calC_\text{WSD}$, we use SemCor \citep{miller1994semcor}. In SemCor each token is annotated with the WordNet sense it evokes. We remove all datapoints with trivial solutions (for which there is only a single sense to choose from, i.e.\ ${|\calD_w|=1}$).

\paragraph{SMD Data} For $\calC_\text{SMD}$, we use the VUAMC (see \cref{sec:md_problems}). 
We remove all part-of-speech types and words not in WordNet (meaning we remove all prepositions, which are almost always labeled as metaphorical in VUAMC, and so are easy to predict). Additionally, because our focus is on conventional metaphors, when investigating SMD performance, we also experiment using a subset of this data in which novel metaphors are filtered out (we refer to this subset as \textbf{Conventional}, as opposed to \textbf{All}). We achieve this using the annotation layer of \citet{do-dinh-etal-2018-weeding}.\footnote{In this data, metaphor novelty is scored with continuous values in the $[-1,1]$ interval; as a threshold, we use $0.2$.} 

\begin{table*}
    \centering
    \begin{tabular}{llccccccccc} \toprule
     && & \multicolumn{3}{c}{\textbf{Relative (ROC-AUC)}} & & \multicolumn{3}{c}{\textbf{Absolute (F1)}} \\
      \multicolumn{2}{l}{\textbf{MPD Model}}  & & \textit{MML} & \textit{OOV} & \textit{Verbs} & &\textit{MML} & \textit{OOV} & \textit{Verbs} \\ \midrule
        \multirow{2}{*}{Ours} & {(with EWISER WSD)}  & & $\hphantom{^{**}}\bm{.78}^{**}$ & $\hphantom{^{*}}\bm{.64}^{*}$ & $.70$ & & $.54$ & $.37$ & $.43$ \\
         &{(with BERT WSD Baseline)}  & & $.72$ & $.60$ & $\hphantom{^{**}}\bm{.71}^{**}$ & & $\hphantom{^{*}}\bm{.60}^{*}$ & $\hphantom{^{*}}\bm{.41}^{*}$ & $\hphantom{^{**}}\bm{.47}^{**}$\\ \midrule
        \multicolumn{2}{l}{MelBERT Average} &  & $.65$ & $.50$ &  $.54$ & & $.54$ & $.33$ & $.37$\\ 
        \multicolumn{2}{l}{Random}  &  & $.54$ & $.51$ & $.51$ & & $.49$ & $.33$ & $.35$ \\
        \multicolumn{2}{l}{Majority}  & & $.50$ & $.50$ & $.50$& & $.00$ & $.00$ & $.00$  \\ \bottomrule
    \end{tabular}
    \caption{Metaphorical polysemy detection results}
    \label{tab:mpd_results}
\end{table*}

\paragraph{Metrics} To evaluate MPD for absolute metaphoricity, we compute the F1-score.
For relative metaphoricity, we compute (for each wordform individually) the area under a receiver operating characteristic curve (ROC-AUC, see \citealp{fawcett2004roc} for discussion). An ROC curve is a plot of the true positive rate against the false positive rate, as a threshold shifts from $0$ to $1$. This tells us whether the metaphoricity probabilities assigned to the senses of a wordform are in correct high--low ordering, even if they are not properly calibrated around the $.5$ threshold. ROC-AUC values range from $1$ (perfectly correct) to $0$ (perfectly incorrect), where $.5$ indicates no correlation. To get a value for a whole evaluation set, we take the mean of the ROC-AUC scores of all wordforms in the set.
To evaluate SMD, we compute the F1-score, and for WSD, where the goal is multi-class classification, we compute the micro-averaged F1-score. 

\paragraph{Significance Testing} We use a two-tailed Monte Carlo permutation test with $r=1{,}000$ permutations. We experiment with two significance levels, $\alpha=0.05$ and~$0.01$, and differentiate between these by indicating the significance level using $^*$ and $^{**}$ respectively.

\subsection{Results} \label{sec:results}

\paragraph{Metaphorical Polysemy Detection} 
MPD results are presented in \cref{tab:mpd_results}. 
We compare two versions of our MPD model, trained in combination with different WSD implementations. 
In all experimental settings (each column), a variant of our model performs significantly better than the highest-performing baseline.
Our model's results are highest for the MML set ($.78$ ROC-AUC and $.60$ F1), perhaps because the examples here exhibited the clearest cases of metaphoricity.
In general, the EWISER WSD variant's results are higher for relative metaphoricity, whilst the baseline WSD variant's results are higher for absolute metaphoricity. The reason for this is unclear. 
Results are lowest for the OOV set, which shows that the model has difficulties generalising beyond training data. 
Generalising to never-before-seen senses is a well-known issue in WSD: \citeauthor{bevilacqua-navigli-2020-breaking} themselves note that their model, EWISER, relies too much on corpus supervision. This problem will affect MPD models if they are trained with our methodology (\cref{sec:model_overview}); to improve on MPD, it will be necessary to collect dedicated training data. 
Nevertheless, our results show that it is possible to learn MPD using existing resources.

\begin{table}
    \centering
    \begin{tabular}{lcc} \toprule
        & \multicolumn{2}{c}{\textbf{F1}} \\ 
         \textbf{SMD Model} & \textit{Conventional} & \textit{All} \\ \midrule
        Ours (WSD+MPD) & $\hphantom{^{**}}\bm{.659}^{**}$ & $.631$  \\ \midrule
        MelBERT & $.626$ & $.638$ \\
        BERT SMD Baseline & $.619$ & $.625$  \\ \bottomrule
    \end{tabular}
    \caption{Standard metaphor detection results}
    \label{tab:md_results}
\end{table}
\paragraph{Standard Metaphor Detection} 
SMD results are presented in \cref{tab:md_results}, for both variants of $\calC_\text{SMD}$ (one with only conventional metaphors, one with any). When trained and evaluated on only conventional metaphors, our combined MPD+WSD model significantly outperforms the state-of-the-art, scoring $.659$ F1 (compared to MelBERT's $.626$).
When all metaphors are included, MelBERT's results are higher that our model ($.638$ compared to $.631$), but in this case the difference is insignificant (at both significance levels).
That our model's performance is diminished in this setting is unsurprising, since this data will contain novel metaphors which are not encoded as senses in WordNet. 
We expect that with improvements to WSD generalisation, it should be possible to improve the SMD performance of a combined MPD+WSD model, for both variants.
As mentioned above, generalising to unseen senses is a common issue in WSD. 
Only $9.4\%$ of words in the SMD (All) test set have all their senses covered in the WSD training data, which is likely to have a substantial effect on the combined model's performance.

\begin{table}
    \centering
    \begin{tabular}{llr} \toprule
         \textbf{WSD Model} & \textbf{Objective}& \textbf{F1} \\ \midrule
        \multirow{2}{*}{EWISER} & WSD & $.768$ \\
         & WSD \& SMD & $.766$ \\ \midrule
        \multirow{2}{*}{BERT WSD Baseline} & WSD & $.740$ \\
         & WSD \& SMD & $.741$ \\ \bottomrule
    \end{tabular}
    \caption{Word sense disambiguation results}
    \label{tab:wsd_results}
\end{table}
\paragraph{Word Sense Disambiguation} 
\cref{tab:wsd_results} compares the performance of WSD models which are paired with MPD models and trained jointly on SMD, as opposed to trained on WSD alone. 
Although the auxiliary SMD objective leads marginal numerical differences ($-0.02$ and $+0.01$ for EWISER and the baseline WSD model respectively), in both cases these differences are not significantly distinguishable (at either significance level), suggesting that utilising SMD data to train WSD models is not useful. This is likely because of the relatively small overlap between the vocabularies of the datasets.

\section{Conclusion and Future Work}\label{sec:discussion}
In this paper, we argued that the standard metaphor detection (SMD) task in NLP is ill-suited to conventional metaphor, because it conflates metaphor detection with word sense disambiguation (WSD). 
As an alternative, we proposed metaphorical polysemy detection (MPD). 
We constructed the first MPD model, which identifies conventional metaphors in WordNet. 
To train it, we employed a novel training technique which exploits resources designed for SMD and WSD. 
To evaluate MPD, we collected two sets of evaluation data, and proposed a new performance measure based on relative judgements of metaphoricity.
Our model performed significantly better than a state-of-the-art SMD model in all MPD evaluation settings (e.g.\ attaining $.78$ ROC-AUC and $.60$ F1 on a subset based on the master metaphor list).
Additionally, we found that pairing our model with a WSD model led to state-of-the-art results for token-based conventional metaphor detection ($.659$ compared to $.626$). We make our code and MPD evaluation data publicly available.\footnote{\url{https://github.com/rowanhm/metaphorical-polysemy-detection}}

Making serious improvements in MPD will require the collection of dedicated training data. 
Training MPD using SMD data necessitated the inclusion of a WSD model, which will negatively affect MPD performance when it makes mistakes.
Additionally, our model did not utilise information about the set of definitions a wordform is associated with ($\calD_w$), and instead made the na\"{i}ve assumption that this information will be encoded in a static embedding of~$w$. 
Set- and graph-based architectures which exploited this information would be a natural thing to explore. 
With more data and better models, it may be possible to synthesise a full set of WordNet metaphoricity annotation, and even to extend this synthetic annotation to multilingual WordNet versions. 
This would be a valuable resource, which would open the door to study other questions surrounding metaphorical polysemy.

\section*{Acknowledgements}

The first author would like to acknowledge the help of Laurence Midgley and Ryan Singh, for conversations about conditionality, and would also like to gratefully acknowledge Magdalene College and the Cambridge Trust, who funded him for the duration of this work.

\bibliography{anthology,custom}
\bibliographystyle{acl_natbib}

\appendix

\section{Annotation Guidelines} \label{sec:guidelines}

\begin{em}
We are studying \textbf{conventional metaphors}. A conventional metaphor is a metaphorical usage of a word that is common enough to appear in dictionaries. Consider the two definitions below for the English verb ``flood''.

\begin{enumerate}
    \item cover with liquid, usually water (e.g.\ ``the swollen river flooded the village'')
    \item supply with an excess of (e.g.\ ``flood the market with tennis shoes'')
\end{enumerate}
The first is a literal meaning, while the second is a conventional metaphor (in the example sentence, the market is not literally being flooded). Words often have more than two meanings, and can have multiple literal meanings and/or multiple metaphorical meanings.

Your task is to identify which definitions are conventional metaphors. To annotate a word:

\begin{enumerate}
    \item Read all of the definitions the word is associated with. Using the definitions (and the synonyms and example sentences), try and get a feel for the meaning of each definition and how they are different. 
    \item Then, for each of the word's definitions:
    \begin{enumerate}
        \item Determine if another of the word's definitions is more “basic” than the current one. Basic definitions tend to be
        \begin{itemize}
            \item more concrete (they describe things which are easier to see, hear, feel, smell, and/or taste);
            \item related to the body or the physical world;
            \item more precise (less vague);
            \item historically older
        \end{itemize}
        Basic meanings are not necessarily the first definitions of a word in the list, and are not necessarily the most frequent meanings of a word.
    \item If the word has a more basic definition than the current one, decide whether the current definition can be understood as an extension of a more basic one.
    \end{enumerate}

\end{enumerate}
If a definition can be understood as an extension of a more basic one belonging to the word, label it as ``metaphorical''. Otherwise, label it as ``literal''. There is one exception:

Sometimes, the definitions can be ambiguous, and combine metaphorical and literal meanings. Sometimes the definition is explicitly ambiguous (it might say that it is meant ``metaphorically or literally'', or that it can apply to something ``concrete or abstract''), and other times it is just vague. If the definition is vague, then you should attempt to resolve the ambiguity by looking at the example sentences. For example, if all the examples are clearly metaphorical, say that this definition is metaphorical.
If it cannot be disambiguated, you should select ``mixed''. 

\end{em}

\section{Additional WSD Models}\label{sec:wsd_models}

\paragraph{WSD Baseline} Suppose we have another $k$-dimensional embedding space for tokens, retrieved by ${\texttt{TokenEmb} : \calT \mapsto {\real}^{k}}$.\footnote{For  $\texttt{TokenEmb}$ we use BERT embeddings; see \cref{app:training}.} As a baseline WSD model, we predict a distribution over $\calS$ from $\texttt{TokenEmb}(t)$:
\begin{align}
& \vh = \texttt{TokenEmb}(t)   \\
& \pwsd (s\mid t) = \texttt{softmax}\big(\texttt{MLP}_\vphi^{\langle k,|\calS|\rangle} \big ( \vh\big ) \big ) \nonumber
\end{align}
In practice, we renormalise the output distribution so only senses with ${d\in\calD_w}$ have nonzero probabilities. 

\paragraph{EWISER} For a high-performing WSD model, we experiment with a reimplementation of EWISER \citep{bevilacqua-navigli-2020-breaking}. Where $O$ is a matrix where the $i$\textsuperscript{th} column corresponds to a $b$-dimensional embedding of ${d_i\in\calD}$,
${\texttt{SynsetEmb}(d_i)}$, and $A$ is an adjacency matrix of size ${|\calD|\times |\calD|}$, where $1$ indicates that two synsets are connected, and $0$ indicates they are disconnected, EWISER is defined as
\begin{align}
& \vh = \texttt{MLP}^{\langle k, k \rangle}_\vphi \big(\texttt{TokenEmb}(t) \big)  \\
& \vz =  \vh^T O \nonumber \\ 
& \pwsd (m\mid t) = \sigma \big(\vz A^T + \vz \big)  \nonumber
\end{align}
In the original paper, a linear layer rather than an MLP is used, and many different experimental settings are explored; we only experiment with the setting in which $S$ is initialised with only hypernyms set to $1$, and $O$ and $S$ are kept frozen. For the definition representations and token representations, we use the same ones as the other models (see \cref{app:training}), and use the standard activation functions we use throughout, rather than the Swish activation function \citep{ramachandran2017swish} which \citeauthor{bevilacqua-navigli-2020-breaking}\ use. In practice, we renormalise the output distribution so only senses with ${d\in\calD_w}$ have nonzero probabilities. 

\section{Additional SMD Models} \label{app:smd_models}

\paragraph{SMD Baseline} We compute a simple SMD baseline, which takes a contextualised embedding and passes it through an MLP to make predictions
\begin{align}
& \vh = \texttt{TokenEmb}(t) \\
& p_\vpsi (m\mid t) = \sigma \big(\texttt{MLP}^{\langle k, 1 \rangle}_\vpsi \big( \vh \big)\big) \nonumber
\end{align}
where $\vpsi$ are sets of parameters.

\paragraph{MelBERT} For a high-performing SMD model, we experiment with a reimplementation of MelBERT \citep{choi-etal-2021-melbert}. Let ${\texttt{SentEmb} : \calT \mapsto {\real}^{k}}$ return a $k$-dimensional sentence embedding (for BERT, this can correspond to the $\texttt{BOS}$ token). We define MelBERT as
\begin{align}
& \vh_\text{tok} = \texttt{TokenEmb}(t)  \\
& \vh_\text{SPV} = \texttt{MLP}^{\langle 2k, k \rangle}_{\vpsi_1} \big( \vh_\text{tok} \circ \texttt{SentEmb}(t)  \big) \nonumber \\
& \vh_\text{MIP} = \texttt{MLP}^{\langle 2k, k \rangle}_{\vpsi_2} \big(\vh_\text{tok} \circ \texttt{TypeEmb}(w) \big)\nonumber \\
& \vh_\text{both} = \vh_\text{MIP} \circ \vh_\text{SPV} \nonumber \\
& p_{\langle \vpsi_1, \vpsi_2, \vpsi_3 \rangle} (m\mid t) = \sigma \big(\texttt{linear}^{\langle 2k, 1 \rangle}_{\vpsi_3} \big( \vh_\text{both} \big)\big)  \nonumber
\end{align}
where $\vpsi_1$, $\vpsi_1$, $\vpsi_3$ are sets of parameters. We also use BERT rather than RoBERTa \citep{liu2019roberta}, for parity with other models.

\section{Implementation and Training Details}\label{app:training}

\paragraph{Embedding Spaces} For $\texttt{TokenEmb}(t)$, we use the output of BERT base \citep{devlin-etal-2019-bert}. Following \citeauthor{bevilacqua-navigli-2020-breaking}, we average the last four layers, and for wordforms which correspond to multiple BERT tokens, we use the first.
For $\texttt{SynsetEmb}(d)$, we average all of the word sense ARES embeddings \citep{scarlini-etal-2020-contexts} associated with $d$ (this approach is also following \citeauthor{bevilacqua-navigli-2020-breaking}), and pass them through SVD (default parameters from \texttt{scikit-learn}, \citealp{pedregosa2011sklearn}) to make them the same dimentionality as BERT.
For $\texttt{TypeEmb}(w)$, we follow \citet{choi-etal-2021-melbert} and compute the BERT embedding where the input is $w$ on its own. In practice, then, the dimentionality of all our embedding spaces are $k=768$.

\paragraph{Data Splits}  Having performed the filtration described in \cref{sec:setup}, and additionally removing any datapoint which does not align with the BERT tokenisation scheme, we finally compute our own datasplits, shown in \cref{tab:train_datasets}. 
\begin{table}[H]
\centering
\begin{tabular}{lrrr} \toprule
\textbf{Dataset} & \textbf{\# Train} & \textbf{\# Dev} & \textbf{\# Test} \\ \midrule
VUAMC (All) & $75{,}395$ & $8{,}818$ & $9{,}594$ \\
VUAMC (Conv.) & $71{,}920$ & $8{,}539$ & $9{,}169$ \\
SemCor & $141{,}025$ & $17{,}701$ & $17{,}481$ \\ \bottomrule
 
\end{tabular}
\caption{SMD and WSD datasplits} \label{tab:train_datasets}
\end{table}
\noindent Because of our data filtering process, numbers shown for EWISER and MelBERT in our paper cannot be compared with the originals.

\begin{table*}
    \centering
    \resizebox{\textwidth}{!}{
    \begin{tabular}{rclcccccccr} \toprule
         \textbf{\#} & \textbf{Experiment} & \textbf{Model} & $\bm{\alpha}$ & $\bm{x}$ & $\bm{n_\phi}$ & $\bm{n_\theta}$ & $\bm{h_\phi}$ & $\bm{h_\theta}$ & $\bm{\gamma}$ & $\bm{\gamma_d}$ \\ \midrule
         $1$ & \multirow{3}{*}{\makecell{SMD\\(All)}} & Ours (w/ WSD Baseline) & $0.8$ & $0.1$ & $1\hphantom{^*}$ & $3\hphantom{^*}$ & $500\hphantom{^*}$ & $300\hphantom{^*}$ & $5\times 10^{-4}$ & $1$ \\ 
         $2$ & & MelBERT & & $0.2$ & $3^\dagger$ & $1^\ddagger$ & $500^\dagger$ & $300^\ddagger$ & $1\times 10^{-4}$ & $10$ \\
         $3$ & & BERT SMD Baseline &  & $0.2$ & $2^*$ & & $300^*$ &  & $5\times 10^{-4}$ & $1$ \\ \midrule
         
         $4$ & \multirow{3}{*}{\makecell{SMD\\(Conv.)}} & Ours (w/ EWISER) & $0.8$ & $0.2$ & $3\hphantom{^*}$ & $4\hphantom{^*}$ & $500\hphantom{^*}$ & $300\hphantom{^*}$ & $1\times 10^{-4}$ & $1$ \\ 
         $5$ & & MelBERT & & $0.1$ & $3^\dagger$ & $3^\ddagger$ & $300^\dagger$ & $500^\ddagger$ & $1\times 10^{-4}$ & $10$ \\
         $6$ & & BERT SMD Baseline & & $0.2$ & $3^*$ &  & $500^*$ &  & $1\times 10^{-4}$ & $1$ \\ \midrule
         
         $7$ & \multirow{2}{*}{\makecell{WSD\\(WSD \& SMD)}} & EWISER & $0.2$ & $0.1$ & $3\hphantom{^*}$ & $4\hphantom{^*}$ & $300\hphantom{^*}$ & $500\hphantom{^*}$ & $1\times 10^{-4}$ & $10$ \\
         $8$ & & Baseline & $0.2$ & $0.1$ & $3\hphantom{^*}$ & $3\hphantom{^*}$ & $300\hphantom{^*}$ & $500\hphantom{^*}$ & $1\times 10^{-4}$ & $10$ \\ \midrule
         
         $9$ & \multirow{2}{*}{\makecell{WSD\\(WSD only)}} & EWISER &  & $0.1$ & $4\hphantom{^*}$ &  & $500\hphantom{^*}$ &  & $1\times 10^{-4}$ & $10$ \\
         $10$ & & Baseline & & $0.1$ & $1\hphantom{^*}$ & & $300\hphantom{^*}$ & & $5\times 10^{-4}$ & $1$ \\ \midrule
         
         $11$ & \multirow{3}{*}{MPD} & Ours (w/ EWISER) & $0.4$ & $0.2$ & $3\hphantom{^*}$ & $4\hphantom{^*}$ & $500\hphantom{^*}$ & $300\hphantom{^*}$ & $1\times 10^{-4}$ & $1$ \\ 
         $12$ & & Ours (w/ WSD Baseline) & $0.2$ & $0.1$ & $1\hphantom{^*}$ & $3\hphantom{^*}$ & $500\hphantom{^*}$ & $300\hphantom{^*}$ & $5\times 10^{-4}$ & $1$ \\
         $13$ & & MelBERT Average & \multicolumn{8}{c}{Uses model $2$ (above)} \\ \bottomrule
    \end{tabular}}
    \caption{Final hyperparameters of the models presented in \cref{sec:evaluation}}
    \label{tab:params}
\end{table*}

\paragraph{Implementation} We implement our models in \texttt{PyTorch} \citep{adam2019pytorch}. Our MLP is implemented so each middle layer is the same size, which is controlled by a hyperparameter. Each layer consists of Dropout \citep{srivastava2014dropout} then a linear layer, then a ReLU activation function (ReLU is excluded from the output of the final linear layer).  
As an optimiser, we use AdamW \citep{loshchilov2017adamw}. We train in batches of $128$ datapoints at a time (from both SMD and WSD datasets simultaneously). 

\paragraph{Training and Loss} In practise, after training the objective in \cref{eq:joint_loss}, we freeze the WSD model, set $\alpha$ to $1$, then continue training to finetune the MPD component. This means that if initially $\alpha=0$, the WSD and MPD subcomponents are trained sequentially. If initially $\alpha=1$, meanwhile, the distribution over $s$ is inferred implicitly, without learning WSD.
More specifically,  every $50$ iterations we compute the loss on the development set, and if this loss does not decrease for $5$ consecutive checks then our early stopping criteria is met. The first time this criteria is met we freeze the WSD component, divide the learning rate by a divisor (a hyperparameter), set $\alpha=1$, recover the best model on the development set so far, and resume training (in effect fine-tuning the MPD component); the second time it is met we recover the best model and stop training completely. 

\section{Hyperparameter Tuning}

\paragraph{Hyperparameter Search} For each value of $\alpha\in\{ 0.0, 0.2, 0.4, 0.6, 0.8, 1.0\} $, we perform a random search over the following hyperparameters: 
\begin{itemize}
    \item The number of layers of $\MLP_\vtheta$ and $\MLP_\vphi$, denoted $n_\theta$ and $n_\phi$, sampled from $\{1, 2, 3, 4\}$.
    \item The dimensionality of the hidden state of $\MLP_\vtheta$ and $\MLP_\vphi$, denoted $h_\theta$ and $h_\phi$, sampled from $\{100, 300, 500\}$.
    \item The Dropout \citep{srivastava2014dropout}, denoted $x$, sampled from $\{0.1, 0.2, 0.3, 0.4\}$.
    \item The learning rate, denoted $\gamma$, sampled from $\{0.005, 0.001, 0.0005, 0.0001\}$.
    \item The learning rate divisor, denoted $\gamma_d$, sampled from $\{1, 10\}$.
\end{itemize}
For each model, and each value of $\alpha$, we train $20$ samples from this hyperparameter space. 

\paragraph{Model Selection} For SMD, we choose models with the best F1 on the VUAMC development set. For WSD, we do the same but on the SemCor set. For MPD, we choose the models with the best mean SMD and WSD performance (again on the development sets).

\paragraph{Final Hyperparameters} The final hyperparameters are shown in \cref{tab:params} ($^*$ are hyperparameters of $\psi$ not $\phi$;
    $^\dagger$ are hyperparameters of $\psi_1$ not $\phi$;
    $^\ddagger$ are hyperparameters of $\psi_2$ not $\theta$).

\end{document}